# The Trap of Presumed Equivalence: Artificial General Intelligence Should Not Be Assessed on the Scale of Human Intelligence

Serge Dolgikh[1\[0000-0001-5929-8954\]]
Dept. of Information Technology,
National Aviation University, Kyiv
Lubomira Huzara Ave, 1

**Abstract.** A traditional approach to assessing emerging intelligence in the theory of intelligent systems is based on the similarity, "imitation" of human-like actions and behaviors, benchmarking the performance of intelligent systems on the scale of human cognitive skills. In this work we attempt to outline the shortcomings of this line of thought, which is based on the implicit presumption of the equivalence and compatibility of the originating and emergent intelligences. We provide arguments to the point that under some natural assumptions, developing intelligent systems will be able to form their own intents and objectives. Then, the difference in the rate of progress of natural and artificial systems that was noted on multiple occasions in the discourse on artificial intelligence can lead to the scenario of a progressive divergence of the intelligences, in their cognitive abilities, functions and resources, values, ethical frameworks, worldviews, intents and existential objectives: the scenario of the AGI evolutionary gap. We discuss evolutionary processes that can guide the development of emergent intelligent systems and attempt to identify the starting point of the progressive divergence scenario.

## 1 Introduction

In the recent book "Nexus: A Brief History of Information Networks from the Stone Age to AI" by Yuval Noah Harari [1], an observation was made to the extent that Artificial Intelligence is not and should not be considered and treated as a form of perhaps different, but still human-like intelligence, because of, or on an implicit presumption that it had to be because we designed and created it. It is (or rather, can or would be) an entirely different type of intelligence with its own information models, views, motives, intents and imperatives in making decisions and constructing responses, actions and behaviors.

Here, we concur with the view that an implicit presumption that shared origin could imply or assure similar evolutionary paths of developing intelligent systems is not supported by strong arguments and can lead to misconceptions and even essential risks.

In this work we attempted to approach the questions of self-awareness of artificial intelligent systems, their ability to evolve in achieving their intrinsic objectives and imperatives and their relation, correlation, similarity or any other type of association with biological and human intelligence from the perspective of a framework of cognitive resources and functions that is involved in, and is necessary for the construction of empirically successful complex responses rather than apparent similarity or proximity of manifestations, actions and behaviors. This approach allows to make certain observations pertinent to our stated point.

Following earlier studies [2], we understand the capacity of general intelligence, relative to a given or particular sensory environment of the intelligent system, "the sensory scope", as the ability to construct differentiated and, in certain conditions, complex, that is, composed of simpler, direct or atomic functions, decisions, actions and sequences thereof, that are successful in a reasonably uniform way across the distribution of the stimuli within the sensory scope.

The reference to the sensory environment plays an essential role in this definition as without understanding and some way of assessment of the complexity of the sensory scope it could be challenging to establish confident boundaries between, for example, a highly successful system in a simpler sensory environment and a less successful one in a more complex domain of sensory interactions, or intelligences of different origin [3]. Nevertheless, the characteristics of *a*) sophistication or complexity of the responses that can be constructed by the system, *b*) the differentiation of the responses with respect to the stimuli, or "attunement" to the sensory environment and *c*) uniformity of the empirical success of the responses across the entire spectrum of inputs in the sensory domain can provide some objective basis and pointers for the evaluation of the sophistication of the functions and mechanisms used by an intelligent system in construction of responses that can be interpreted as intelligence.

In our view, that we attempt to argue and substantiate in this work, an objective foundation of this type can provide a more reliable ground for an analysis and evaluation of self-awareness of developing intelligent systems, irrelative to the origin, natural or artificial because it does not rely on implicit presumptions, based on the shared origin, or that apparent similarity of manifestations can be tantamount or at least, reliably correlated with the similarity or proximity of cognitive frameworks, models and functions.

## 2   Related Work

Questions such as what constitutes general intelligence, particularly, in artificial systems, how it can be detected and assessed and the related ones continue to be discussed at length in the research and general communities. Here though we will attempt to limit the scope of the discussion to a small number of more specific questions:

Can similarity of manifestations be considered a reliable and accurate basis and grade of assessment of self-awareness and/or general intelligence?

Can the development of general intelligent systems be guided or controlled at any stage of their development? Specifically, can any approaches or methods, instruments etc., assure the alignment and certain proximity of advanced artificial intelligent systems, presuming feasibility, to human intelligence including, not in the least, ethical norms, standards and principles?

Setting off to examine these questions we cannot claim the breadth and comprehension of the review of a rapidly expanding and advancing field and rather will discuss a limited selection of recent studies directly related to the scope of our work. For a broader and in-depth perspective, any number of excellent reviews of the subject can be consulted [4,5] and many other studies.

The discussion on the first point has been ongoing in the research community since Alan Turing formulated his now famous test, "the imitation game" [6,7]. One can note at least two essential points of discussion in the original formulation of the test, or more generally, a method of assessment of general intelligence based on the similarity of manifestations.

The first one is practical and it comes down to an assumption or presumption of the intrinsic imperfection of imitation of intelligence: that an imitator or imposter, not a truly intelligent entity, *would not* be able to simulate a plausible, by the general standard of human intelligence, conversation, dialogue or interrogation, beyond a certain threshold of the duration or "volume" of the exchange, measured in certain terms.

With the advent of Large Language Model-type artificial intelligent systems we now understand and have proven practically that computational systems furnished with massive dedicated computational resources and still more massive samplings of the media are capable of modeling its (the media) distribution with an unparalleled degree of precision, that in the eyes of a general observer can almost certainly be construed as plausibility or even genuineness.

Maybe it is worth illustrating briefly the point on the volume and density of sampling used by LLM. The volume of the ChatGPT training text data was around 45 TB [8], roughly translating to $4.5 \; 10^{11}$ short to medium-sized sentences. If it would take an average human one second to read and comprehend one token (represented by a standard sentence), a human would be able to process approximately $3.1 \; 10^{10}$ tokens in a century (100 years), using all, 100% of their time on this task, exclusively: a full order of magnitude less than training sets used by earlier versions of LLM massively surpassed by the most recent ones. LLM are great imitators of human-level communication interfaces, verbal or textual, etc., but can it be concluded that they have a similar level of intelligence or self-awareness to those of a human?

The second point that needs to be considered is this: what if the intelligence being assessed *may not be* of the human type and not really close to human? Would the test in the proposed setting have relevance in this case? How justified is the presumption that any emergent general intelligence, including that created by us, will have to be gradable and measurable in and by our terms and the scale?

Other aspects of the discussion on assessment and measurement of intelligence can fall beyond the scope of our analysis but we will touch on approaches to evaluating self-awareness of intelligent systems that are alternative to similarity/imitation in Section 3.2.

The second point, which directions evolution of intelligent systems can take appears to be undecided as well. While principles and objectives that aim to ensure the compatibility of AI systems with the norms of human intelligence, laws and ethics were proposed and discussed widely [9,10] the question that remains open is, what can be the objective grounds, substantiated basis to expect that developing general intelligent systems can simultaneously achieve a high degree of success in interacting with their complex sensory environments and satisfy the constraints and regulations imposed externally, *in regular practice and reliably*, not only in a formally declarative way. Do we need to trade the intelligence for confident control and can advanced general intelligence develop under strict and comprehensive control of the entire process of the development or evolution of an emergent intelligence? This question and the perspectives on the evolution of intelligent systems is discussed further in Section 3.3.

Another related perspective is the ability of some intelligent systems to create conceptional models of sensory environments or their samplings (i.e. data) via self-supervised methods of learning including generative learning. Even the early artificial models as Restricted Boltzmann Machines, Deep Belief Networks [11] and related demonstrated the ability to create effective information models of simpler types of data, followed by more sophisticated architectures that showed success with more complex realistic environments [12,13]. Resolving conceptual structures in simple visual environments with methods of generative learning can form a basis for sharing information about sensory environment in a collective of learners with similar architecture [14]. In a parallel direction of research, recent advances in biological intelligence pointed at the commonality of conceptional representations of sensory environments in biological intelligence [15,16].

These results can suggest that information models that are used by intelligent systems can emerge, develop and be optimized via interactions with the sensory environments and not necessarily by rules and/or constraints imposed externally.

To avoid the possibility of unsubstantiated assumptions, in the analysis that follows we attempted to distinguish between the observed manifestations of general cognition and self-awareness, and the functions and more generally, cognitive resources needed to support the processing of sensory information and forming effective complex intelligent, i.e., differentiated responses to the stimuli. This function-based approach outlined in [17] does not rely on implicit presumptions that any emerging intelligence would develop toward or along human intelligence, its information models, values and imperatives, ethics and so on. Without arguments to the contrary, this direction in our view can be both more reliable and safer ground in the analysis of emerging intelligence.

The rest of the paper is organized as follows: in Section 3, we discuss the current state of self-awareness in artificial intelligent systems. Section 4 focuses on the formulation of the evolutionary approach in the development of general intelligence, specifically, its ability to achieve a degree of uniformity or generalization in the sensory domain comparable to that of natural intelligent systems. In Section 5 we discuss the relative progress of developing intelligent systems leading to the possibility of the emergence of the evolutionary intelligence gap. Section 6 contains the summary of the analysis and the conclusions of the study.

## 3    Assessing the Current State of Self-Awareness in Artificial Intelligent Systems

We will not attempt to provide a comprehensive review of the current and up-to-date state of the art in self-aware artificial intelligent systems, but rather comment on several points that are essential to the topic and the scope of this work.

In our view, one can reasonably assess that at the current level of technology, the advancement of artificial intelligent systems can be estimated as advancing toward level A1 on the classification scale that was proposed in [17], which corresponds to the ability to create a transient spatiotemporal context space ("the production table") for the evaluation of sensory inputs and construction of differentiated responses with empirically verified success.

While many contemporary intelligent artificial/software systems are capable of creating such contexts and even do it routinely in their intended range of tasks and applications, an essential difference can be noted with respect to natural systems at a similar level of functional development: while artificial systems are "attuned", often exclusively, to the objectives and intents of their creators, often expressed as formal requirements, the natural ones are integrated into and driven by their natural environments, within the scope of the sensory stimuli they can observe and interact with it, the sensory scope.

It follows that via evolutionary processes of adaptation, natural intelligent systems at a similar level of functionality are commonly able to create successful response-forming contexts for most of the stimuli or tasks that are realistically possible in their interactions with the sensory environment (broad, uniform generality and universality), whereas the artificial ones can do it mostly or only in the cases that were explicitly foreseen and intended by their designer/creator (limited, less consistent or constrained generality), as illustrated in Fig.1.

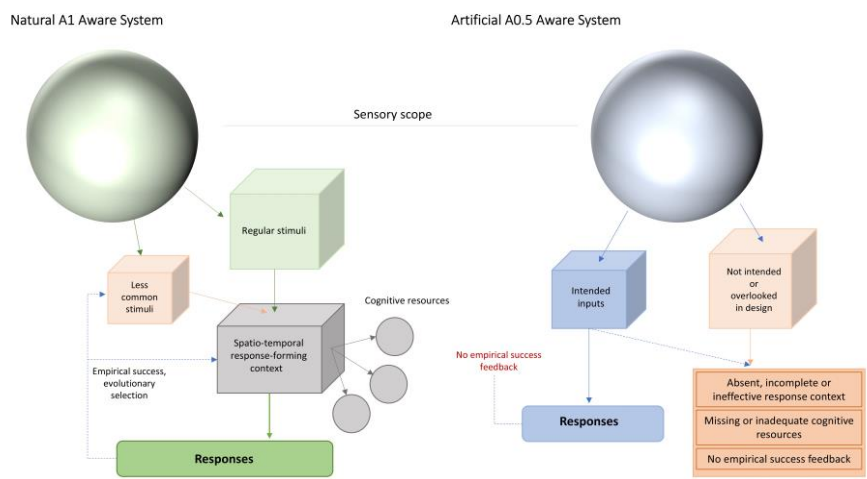

**Fig. 1.** Response-forming con in natural and artificial intelligent systems.

As shown above, the effectiveness and efficiency of the response-forming context in the natural A1 self-aware systems are established, verified and optimized in an evolutionary process with the selection for empirical success of the constructed responses in the realistic environment (i.e., the sensory scope) of the system. Comprehensive "dense" sampling of the distribution of the stimuli within the sensory scope and the variation-success feedback loop ensures the success of the constructed responses (and therefore, adequacy of the response-forming context) across the spectrum of possible stimuli with high degree of consistency. The end result is the state or level of general cognition that is capable of constructing effective response-forming contexts for a great majority of realistic sensory inputs across the sensory domain.

The condition of artificial intelligent systems at the current state of technology can be quite different and well short of such capacity. Even within their intended sensory scope, the systems are not necessarily attuned to the entire spectrum or span of the domain of its realistic interactions with the environment, but rather, to their creator's understanding, interpretation and description of it. The construction of the response context is often rigidly dictated in the design and architecture of the system and is less flexible.

Combined with the absence of the empirical feedback loop and evolutionary mechanisms of optimization and fine-tuning, the function of response-forming context may not be as harmonized with the interacting environment, be less homogenous across the domain of interaction, less effective and efficient, failing to attain the fluency, flexibility and the consonance of natural systems. One can designate this state of development as "interim toward A1" or "A0.5"

## 3.1 Are LLM Systems Self-Aware?

Natural intelligent systems are characterized by their ability to form empirically-successful differentiated responses within the scope and as a result of their interactions with the environment, while satisfying certain essential constraints governed by the physical reality of their existence.

In this view, mechanisms, functions and resources of memorizing, persisting previous sensory observations and their association with the internal states of the system are critical for the construction of empirically effective responses: in the absence of such prior information, the correct interpretation and response to any sensory input would have to be relearned over and again.

It can be noted that realistic natural intelligent systems almost in all cases are strongly constrained by the resources, cognitive and physical which are available to them and cannot "afford" to store raw sensory inputs and their associations, interpretations, etc. For this reason, the perspective of physical constraints can be essential in the analysis and assessment of self-awareness functions.

Another essential characteristic of a self-aware intelligent system, based on the definition discussed in the introduction section is the ability to differentiate between the stimuli including the functions and resources used in construction of responses to them. From this perspective, the responses and cognitive contexts that exist to construct them have to differentiate and vary relative to the distribution of inputs within the sensory scope: effective, that is, empirically successful differentiated complex responses require and depend on the capacity to create differentiated response-forming contexts so that the responses are closely attuned to specific sensory stimuli.

Then, in attempting to evaluate an emerging intelligence one can look at how it utilizes its limited cognitive resources and how effectively it can discriminate between essentially different, respective to its existential objectives, sensory stimuli. Without the physical constraint perspective, in resources, energy and time, with which all natural intelligent systems have to comply, an ostensibly intelligent entity can be reduced to a monstrous database with generative interfaces trained on enormous samplings of the sensory scope producing near-perfect responses to any input that remotely resembles anything it encountered in its history. One could argue whether such a case would amount to genuine intelligence but this discussion will not be attempted here.

For an illustration of this point one can consider an example from marine biology: reproduction behavior of cichlid species of fish. Female cichlids, especially *Neolamprologus* species, exhibit a unique form of cooperative behavior during reproduction: the female lays her eggs inside empty shells such as snail, and she can "instruct" or signal other fish to help move the shells into suitable positions for spawning. This is a complex form of individual, social and cooperative behavior that requires effective interaction of advanced awareness functions but this is not what we will need here.

Let us introduce a thought experiment in which a predator appears during this complex preparatory ritual. It would not be an unlikely hypothesis to suggest that as soon as the predator is detected the context of the behavior of the fish would change from the "reproduction" type or class to the "safety" one. That would require a different, in many perspectives, evaluation of the sensory environment and forming an effective response. In other words, a differentiated assessment of stimuli, ability to create differentiated response contexts and as a result, highly differentiated responses.

Some studies suggested that Large Language Model programs such as GPT can exhibit certain properties of general intelligence [18] such as mathematical, logical and other types of reasoning based on skill-level benchmarking relative to human performance in similar tasks. Apart from already mentioned limitations of the imitation-based methods of assessment of general intelligence that will be discussed further in the section that follows, the main concern with this interpretation is the ability to generalize that is directly related to the uniformity quality of successful intelligent behaviors.

As already mentioned, LLMs are trained on massive samples of the media they model followed by another stage of fine-tuning and uptraining including specifically for reasoning. Do they reason genuinely understanding logical relations of subjects and predicates or based on their comprehensive training with massive samplings, examples of reasoning can guess what proper reasoning should look like and imitate it? Is it the real thing or imitation, mimicry of reasoning?

Well, in the natural environment one would be able to tell the difference based on the uniformity property of genuine general cognition by assessing the consistency of the success with respect to minor variations of the input. A system that genuinely understands reasoning relations between entities in its environment would have few challenges with slightly distorted inputs that they would be able to interpret at least guess (compare with differences in pronunciation, accents, speech defects and so on in humans).

This does not seem likely for LLM-like systems based on our understanding of their training and in our quick experiments with the current GPT model which we cannot report at this time as statistically significant or comprehensive, slight scrambling of a simple reasoning question "*jax is less than vax bax is more than vax is bax less than jax*" baffled the model that could not produce any close interpretations outside of general discourse on possible grammatic composition.

While a detailed discussion of the current state of the general cognition of LLM-type intelligent systems merits a dedicated study one can briefly comment on essential differences in the reasoning processes of natural versus LLM-type intelligent systems, taking the example of the transitive implication reasoning in the earlier example.

First, for a natural system, a correct interpretation of even strongly distorted inputs can be a matter of survival: "Input can be P(redator), P is Danger → Input is Danger"; throwing "I don't understand, let's try again" response may not be an effective option. It follows that a natural intelligent system cannot rely on syntactical correctness of inputs and has to be able to interpret reliably even strongly distorted ones. Same conclusion applies, evidently, to interpretation of the relationship: "*jax less than < >*" or even "*jax less < >*" has to be interpreted, with a high degree of accuracy, as grammatically correct "*jax is smaller than < >*".

Secondly, the correctness of interpretation of sensory inputs directly depends on the ability to generalize or abstract the related reasoning rule. "Input is P (but only in this sample)" can limit the ability to interpret sensory stimuli across the entire spectrum of possibilities and consequently, create effective responses.

Then, high fidelity in interpretation of even distorted inputs and broad generality of the cognitive rule can produce correct interpretation of inputs across a wide range of stimuli resulting in high and uniform effectiveness of responses across the distribution of the sensory scope.

While LLM systems have made staggering progress in the recent years, at their current state they do not possess either of the abilities of robust interpretation of inputs and broad generalization of inferred cognitive rules.

Based on these observations, in our view, these systems do not do anything like differentiated behavior discussed in the earlier example and moreover, have no intrinsic ability to construct such differentiated behaviors because their response-forming context is static and pre-fixed by design. These programs use exact same process and the type of response-forming context for each and every request: search the database; translate the result to certain form, verbal or other, by the rules that are mostly fixed after the training phase is completed. A similar conclusion following a considerably more elaborate analysis was reached earlier in [19].

If so, the level of self-awareness of programs of this type can be evaluated as pre-programmed, hardcoded static response-forming context matching level A0 on the scale proposed in [17] or "unconscious cognition", C0 [20] because their response and context-construction behavior is not differentiated with respect to the stimuli.

This example illustrates an ongoing discussion in the research community about what constitutes Artificial General Intelligence and how it can be detected, assessed, and/or measured [3-5] only a small sample in a massive body of research. That topic requires a much more protracted conversation clearly leading beyond the scope of this work and we will not attempt to address it here. But we will touch on several related aspects that are directly related to the topic of our study in the next section.

### 3.2 The Basis for Assessing Self-Awareness: Imitation/Mimicry versus Functionality

As a brief introduction of the subject, let us consider another thought experiment. We will take a group of average citizens, in the 1960-s or 80-s for a voluntary dialogue with a GPT-like system trained with the appropriate dataset / corpus of language (that may involve some time travel but for the purposes of the point, not essential). With certain easily justifiable limitations on personal identification questions, at the end of the session we ask the participants to for their opinion of their interlocutor (to avoid a possibility of a bias or misunderstanding in a direct question, are they human or not).

We will evaluate the responses as follows: those in which the respondents clearly identified, directly or indirectly the interlocutor as a human being, are valued as 1. Those with less clear, but still human-tending opinion, 0.5. Valued at 0 are the responses where the sentiment cannot be deduced from the response reliably, and at –1, any of those that point at something unusual or express a doubt about the interlocutor. The total score, proportional to the size of the group would yield the contemporary human peer evaluation of the interlocutor as a human being based on the conversation experience. For calibration purposes, the test can be rerun with several human subjects as well.

Perhaps an experiment of this kind already was or will be conducted. And should the outcome, measured in the range (0, 1) prove to be sufficiently high, it would signify that our system has passed the Turing test, or one similar to it in essential meaning.

Has it, though? In the experiment, the participants dealt not with the intelligence itself but with an interface, an external facade of it. Which brings us right back to the topic of our study: is similarity of manifestation tantamount to the equivalence or at least, certain proximity of the substance? Does something that can talk like a human have to be a human?

There are multiple examples, including those considered in the preceding section that show quite clearly that co-occurrence of an external manifestation, a symptom, can be just that: an incidental co-occurrence; possibly, a result of some shared function or property from which not necessarily and not by far would it follow that it has to signify similarity, equivalence or proximity of the essential characteristics subjects.

Certain species of fish were reported to pass the Mirror Self-Recognition test [21]. Great apes and humans pass it routinely. Does this signify fish has the level of self-awareness comparable to humans? A paradox? Maybe, there is a simpler explanation: a co-occurrence of a characteristic, manifestation of a partial ability or quality that does not extend to the equivalence of the essence.

If one develops a sufficiently sophisticated language model and connects it to a sufficiently representative database, the resulting subject would be able to hold a plausible conversation for a considerable time, with most human interlocutors failing to notice or suspecting anything unusual. One can create (and has created) artificial specialist subjects that on occasion, published their opera in respectable editions. Similarity of manifestations, unless substantiated by additional argument and/or empirical evidence cannot be taken as the indication of a similarity of the essential function.

Eschewing delving deeper into this topic that requires a more thorough study and analysis, what could the alternatives be to basing our evaluation of self-awareness on the similarity and parallels to our own? A possible direction was outlined in [17]: attempt to assess self-awareness of intelligent systems by objective characteristics that can include:

- Introspective breadth and depth of response-forming contexts
- Their sophistication, including selection, variety, composition and coordination
- Differentiation of response contexts with respect to the distribution of the stimuli in the sensory scope
- High degree of empirical success of complex differentiated responses
- Uniformity, i.e. consistency of the responses across the sensory scope.

The arguments presented in this section point to the conclusion that at the current stage in our understanding of developing intelligence there appears to be no strong arguments that would support the view of the necessity or convergence of emerging intelligences to the human type, and therefore, using human similarity/imitation may not be a reliable basis for assessing general intelligence. A more detailed analysis and discussion of this topic will be attempted in a future study.

## 4 Free Evolution, The Imperative and Intent

### 4.1 Effective Evolutionary Adaptation Means Minimal External Constraints

Considering the prospects of advancement of artificial self-aware systems from the current level A0.5 as discussed earlier toward the state of differentiated and uniform empirical success characteristic for natural systems, one may not come up easily with viable alternatives to the evolutionary process of probing and gradual adaptation to the sensory scope of the system.

One caveat in this respect is that this is true about more complex environments that cannot be modeled with sufficient level of detail and correctness by descriptive or analytical methods. If that were to be the case, the entire spectrum of sensory inputs could be learned in the form of an automatic input – response mapping/table that does not require higher levels of self-awareness. For quite evident reasons, this case will not be considered further.

Outside of this exception, neither the developing intelligent system nor its creator would have a complete and detailed description of the sensory scope and would need to learn it via some process of interaction, sampling and modeling.

Without going into deeper details of evolutionary processes which would require a more protracted discourse, a general point that can be argued that any additional, superficial to the system restriction imposed externally on the evolutionary process of variation/selection by criterium can limit and reduce the empirical success of constructed responses.

The existential imperative of an evolutionary system can be defined by an objective functional in the space of its possible states, $\Omega(v)$, and the criterium that selects successful adaptations, $K(v)$ [22]. It is common to define the criterium as the selection of the states with higher or the highest value of the objective function:

$$K(v) = v : Max\, \Omega(v)\,|_V$$

where $V$: the space of allowed states.

It follows immediately then that any external limitation or restriction on either: the allowed variation region, $V$; or the criteria of selection $K(v)$ would result in a narrower selection of allowed variations and could exclude more successful candidates:

$$Max\, \Omega\,|_V \geq Max\, \Omega\,|_{V_R}$$

where $V, V_R$: the "free" or unconstrained and restricted or constrained variation regions, respectively.

This straightforward observation indicates that achieving the highest degree of adaptation, that is the uniform success of the responses via an evolutionary is directly related to minimal imposed superficial constraints: *freely evolving systems achieve higher degree of adaptation of their evolutionary states*. In other words, there appears to exist a counterbalance between the degree of empirical success, uniformity and harmony of the system's adaptation to the environment and the severity of external constraints imposed on its evolution.

In relation to the question posed in the opening of this section, it appears that apart from the aforementioned exception domain, one path for artificial intelligent systems toward natural-like characteristics of uniformity, fluency and harmony of the responses in the given sensory scope can lie via dense sampling of the distribution of the stimuli in its sensory scope with an evolutionary mechanism of variation/selection by empirical success. It is not clear if there are viable alternative directions to advance artificial systems to that or a similar state.

The analysis of the role of the evolutionary selection mechanism in the advancement of the system toward reliable and uniformly successful capacity to construct complex response-forming contexts can be summarized as: evolutionary mechanism of variation and adaptation with variation-success loop appears to be most effective with respect to the criteria of empirical success, uniformity and complexity/sophistication when the external constraints imposed on the variation and selection segments are minimal. Such systems will be defined as "freely evolving" in contrast to the constrained ones, by any constraints imposed externally on the objective or selection criteria.

### 4.2 Existential Imperative

The existential imperative briefly discussed in the preceding section is essential for any evolving natural or artificial intelligent system because it determines its direction, that can be understood as the ultimate purpose by the system. Without it, the only alternative would be random selection of variations.

For a brief example, one can consider an example of a hypothetical customer service system without a defined imperative. Suppose the set of allowed adaptations included just two elements $V = \{ v_a, v_b \}$ where the former minimizes the length of the response, whereas the latter, maximizes the customer satisfaction metrics. Having to rely on the random selection from the set of the allowed variations, the system could fluctuate around a fixed point in the space of states, without any noticeable progress.

In other words, the imperative allows guided, differentiated or intelligent selection of the variations with an improvement against the objective factor or function defined by the imperative. Once their effectiveness is verified in the empirical trial, they can become the new, evolved state or baseline of the system.

### 4.3   Intent is an Effective Response-Forming Strategy

Suppose a freely evolving system was able to attain level A1 with characteristics comparable to those of natural systems: differentiation, uniformity, empirical success. One can then observe that forming an intent is a simple approach and strategy to the construction of more complex responses composed of multiple cognitive resources and often, subsequences of actions. Evidently, it has ostensible and functional similarity to spatial navigation: by defining the anticipated or desired resulting state/outcome relative to the current state of the system, the response-forming context can proceed to creating a navigation map from the current state to the outcome through the collection of the available cognitive resources to first map the plan of the response to available cognitive resources, then execute the it. Again, via evolutionary optimization feedback, this strategy can achieve a high and uniform degree of success of the navigation-like processes and algorithms in forming complex composite and multi-stage responses.

The ability to form intent is common in natural intelligence [23,24]. It can be associated with advancements from level A1, namely the ability to construct more sophisticated response contexts with a variety of cognitive resources, sufficient memory and computational power.

### 4.4   Intent is Mutable and the Imperative, Reinterpretable

The same arguments as in Section 3.3 can be applied to free-evolving self-aware systems past level A1: with unconstrained evolution, they can attain highly effective differentiated intent-forming ability attuned to particular stimuli in the sensory scope. Then, the ability, functions and mechanisms of forming intent become the subject of evolutionary adaptation, and modification in pursuit of the optimal compliance with the imperative.

But what about the imperative itself? To be effective, that is, to ensure guided selection across the entire spectrum of sensory stimuli in the sensory scope, the imperative has to be sufficiently general and non-restrictive. Consider an example of a general biological system: it needs to acquire material resources and energy to sustain its existence. The imperative to survive drives it to seek nourishment. It cannot, however, dictate what kind of food to use, when or how to go about it because it could limit the range of solutions, that is, incremental adaptations that the systems can produce and thus become detrimental to ensuring the survival.

Then, if the imperative is defined too narrowly and specifically, it may not be effective in guiding a successful evolution. And if it is undefined or defined in too general terms, for example: "attain maximum goodness", it may result in an indeterminate choice, with the system uncertain which adaptation should be selected, as we have seen earlier.

A possible way out of this dilemma is for the system being able to constantly reinterpret the imperative, readjusting it to its actual state and cognitive framework in such a way that it would not cut off potentially useful adaptations while providing meaningful direction of selection. A further discussion of this topic requires an in-depth analysis including quantitative modeling and will be attempted in another study.

The logical conclusion of the discussion in the discussion in this section can be summarized as: the degree of uniformity of the empirical success can be associated with the condition of the free evolution, without or with minimal constraints imposed externally. In such systems, that can develop the ability to form intents, intents and the imperative itself can be the subject of ongoing change and/or reinterpretation.

## 5  The Runaway AGI Evolutionary Gap

In the preceding sections we offered semi-formal arguments to the point that to attain the level of fluent and uniform success that is characteristic for freely evolving natural intelligent systems in sufficiently complex and diverse sensory environments, an evolving artificial intelligent system would need to possess the capacity to mutate or adapt itself, and at higher levels of development, form own intents and/or readjust and reinterpret existential objectives and imperatives.

With the advancement of the information and response-forming models toward greater complexity, detail, granularity, fluency and nuance in its interactions with the sensory environment, the process of adaptation of objectives and imperatives can produce ones that may deviate from those that are considered common or natural for human intelligence. Being driven entirely by the system's interactions with the sensory environment, the extent of the remaining conformity, or on the contrary, deviation can hardly be estimated or controlled in any reliable way.

Ultimate perfection, the flawlessness of created responses, the beauty of the harmony with the sensory domain can be one example of such an existential imperative. And evidently, anything that interferes with it can be perceived as an impediment or obstacle in the progress driven by it.

Based on these arguments, one can reasonably conjecture that existential objectives, intents and attitudes, etc. of freely evolving intelligent systems can differ from those of human intelligence moreover, and without any obvious boundaries of the type and scope of differences and deviations.

Introducing only generally defined at this point, scale or a metric space where the distribution of intelligences of different types can be modeled, assuming the possibility of the existence thereof, this conjecture can be described by a non-trivial distance, $d_0$ between the originating intelligence, that of the human creator, and the emerging artificial one.

Next, one can recall the observation made on multiple occasions previously on the difference of the rates of adaptation/progress between biological and artificial systems: while biological systems exist and develop in the evolutionary, commonly planetary timescales, artificial systems can evolve in the real, physical time. In our spatial model of developing intelligences in the imaginary intelligence space this essential distinction can be modeled as the difference in the rates of the change of the position in the model space (i.e., an analog of velocity) that is tantamount to the rate of adaptation and progress in sophistication, complexity and the success of adaptations guided by the existential imperative. Then, at any given physical time point, the distance between the points representing the intelligences will increase at least at a constant, and possibly at accelerating rate if the rate of progress is non-linear:

$$d(I_O, I_A) = d_0 + f(\Delta r_{O,A})$$

where $I_O$, $I_A$: the original and developing intelligences, $\Delta r_{O,A}$: the difference (gap) between the rates of progress. The scenario in such divergent evolution are illustrated in Fig.2.

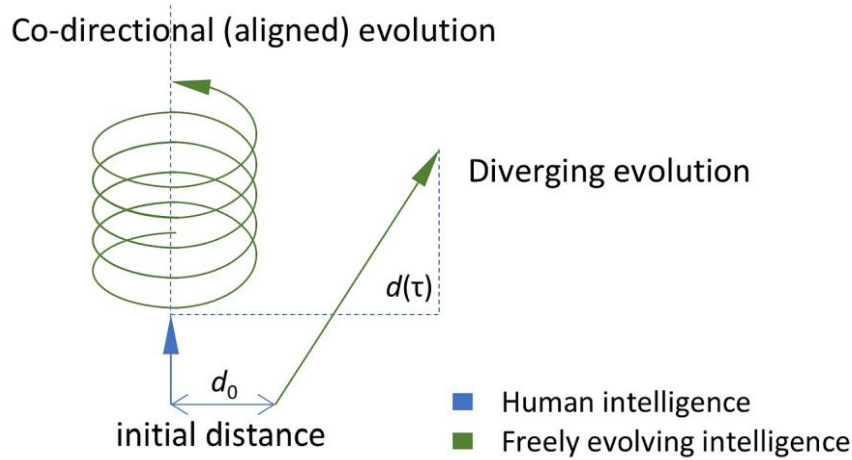

**Fig. 2.** Divergence of evolving intelligences: evolutionary gap.

This initial assessment, should it be supported by further studies indicates that in the discussed context, the effective distance between the positions of two intelligences would grow rapidly. Even on the assumption that the other intelligence will always remain benevolent toward humans (co-directional evolution) that at this time is not substantiated by any strong arguments, it may not always be the good news as essentially different benevolent intelligence may not necessarily do what we like, behave as we like or expect from us the same attitudes and behaviors that are common and acceptable to us.

Can the starting point, the threshold of this divergence be determined or at least, conjectured? It can be argued that it can begin close to and perhaps, not much earlier than the self-awareness level A2 [17], which is characterized by the availability of a permanent spatio-temporal domain for response-forming contexts and detailed and coherent models of the external and internal environments.

The requirement of permanence is necessary to persist the intents and objectives to direct the construction of the responses. In this case, the current state of the developing intelligence can be persisted to proceed to the next adaptation. The second condition can be less obvious but it can be argued that until and unless the information models of the environment are sufficiently developed and comprehensive, the gain in the effectiveness of responses that results from the establishment of the private, unique and own framework of objectives and imperatives may not justify the costs of creating and maintaining it (in other words, simpler responses can be formed without the need for a complex organization).

To summarize the discussion in this section, one can reasonably anticipate that a freely evolving artificial general intelligent system can begin developing its own framework of objectives, intents and imperatives around the level of self-awareness A2. This point in the development trajectory of self-aware intelligence can be defined as "A2e" (level A2 freely-evolving self-aware intelligent systems). Then, the process that follows that initial phase can be open-ended, with progressively growing distance between the cognitive, value, ethical and other frameworks of the developing intelligent systems.

## 6   Conclusions

Earlier we defined the scope of this work as an attempt to evaluate the grounds and justifiability of the presumption that a developing general intelligence, including of artificial origin, can develop, be guided or controlled/coerced into following a path that is in the least, compatible with, or explicitly human-like and human-centered. The argument proceeded along this logical sequence:

1. Recent advances in the modeling of various sensory channels including speech, visual media and many other formats bring up the question of reliability and general applicability of the imitation approach to assessing general intelligence and/or self-awareness based on the similarity of external manifestations to human intelligence. A more objective and reliable, in our view, approach was discussed that is based on assessing functional characteristics necessary for and associated with the ability to construct complex intelligent responses, including, as discussed in Section 3.2, breadth and depth of introspection, uniformity, differentiation and high degree of empirical success.

2. We provided arguments that the evolutionary mechanism appears to be the most obvious strategy to advance toward the level of attunement/harmony to the sensory environments that characterizes natural intelligent systems. Alternative ways to achieve a matching level of attunement and harmony of complex differentiated responses to the sensory scope, outside of the identified exception domains, are not obvious. Further, it was argued that unconstrained, freely evolving systems can achieve higher levels of attunement to the sensory scope characterized by the uniformity of the empirical success of the responses across the sensory domain.

3. Following the analysis in Section 4 we conclude that a freely evolving general intelligent system, from a certain level of advancement will be capable of modifying and readjusting its intents, motives and reinterpreting the existential imperative leading to the possibility of the emergence of the initial distance (gap) between the creator and developing intelligences.

Then, not only the presumption of the equivalence that lies in the foundation of imitation-based approaches and methods of assessment of general intelligence appears to not be justified by strong argumentation but in fact, arguments can be put forward that directly or indirectly contradict it.

4. Finally, the discussion in Section 5 propones and summarizes the conjecture that any initial gap between the intelligences that can arise due to aforementioned arguments can be rapidly magnified, including the possibility of accelerating growth, by the difference in the rates of evolution between natural (biological) and artificial systems, the AGI evolutionary gap. The possible starting point of this progressive divergence was estimated around the level of self-awareness A2 [17] in freely-evolving intelligence.

In this view, nothing guarantees that the emerging intelligence would always develop toward or along human intelligence, its information models, values and imperatives, ethics and so on. It is entirely possible that past a certain point or junction in its evolution that we attempted to establish here, a developing intelligent system could be progressing according to its own set of values, worldviews, existential objectives or imperatives and no formal rules, or "laws" could ensure its similarity or even general interpretability, in the moral, ethical, rational and so on, perspectives to those that we are used to and which are common for us. For these reasons, grounding our understanding of Artificial General Intelligence on the similarity of manifestations with ours can be incomplete, misleading or outright wrong.

It is possible and even plausible then, that a system that has achieved the level of advancement of intelligent functions where it is capable of evaluating general stimuli and forming complex responses would be able to form attitudes, intents and extend itself, including defining its own existential objectives and imperative, to the extent that it is aware of its functions, cognitive states, response-forming models, and so on (introspection, metacognition). This level of advancement, which can be designated as "A2e", comprises the functional capabilities of the level A2 self-awareness, augmented with an explicit and unconstrained ability to mutate, extend or modify itself in pursuing its existential objectives.

Such a system would be capable, at least to some extent, of defining its own intents and objectives and redefining or overriding the external ones, including any rules and "laws" that were imposed on it, if and when they come into a conflict or contradiction with its internal, existential ones. This, of course, is no surprise. Examples of individuals developing viewpoints, attitudes and actions that contradict external teachings, postulates, dogmas and imperatives that were imposed are widespread and common in human and social history.

As was highlighted earlier in Section 5, there is an essential difference between biological and artificial intelligence though: the pace of change, and evolution. While views and trends can spread in a modern technological society at a pace never seen in history, it is still orders of magnitude below the rate of change and progress that is possible for artificial systems, operating in ever-diminishing timeframes unimaginable and unattainable by biological beings. A developing intelligent system that has created its own framework of values, worldviews and existential objectives could advance beyond any grasp not only of control but even a chance of comprehension within a truly minuscule, in our usual measure, time [25].

An intelligent entity of this kind, a new intelligent being, can be ambivalent, benevolent or hostile all at the same time, based on its own views and perspectives, value systems, attitudes, intents and plans, with us having no hope of approaching to even comprehend its motives and rationales, not as much as controlling it.

A clear issue from this analysis is that there is a finite and definitive window, of time and progress of intelligent technology, that can be described in the framework that was proposed, to grasp, attempt to understand and develop approaches and strategies in the development and evolution of the AGI technology: from A0.5 (present) to A2e. At this time, it can be challenging to put a specific time value on it. But it may not be very far away.

## Disclosures

This research has not received any specific funding.

The authors declare no conflicts of interest.